\documentclass[a4paper,conference]{IEEEtran}
\IEEEoverridecommandlockouts

\usepackage{xcolor}

\newcommand{\thename}{Light3DPose}

\newcommand{\R}{\mathcal{R}}
\newcommand{\U}{\mathcal{U}}
\newcommand{\V}{\mathcal{V}}

\usepackage{ifpdf}

\ifCLASSINFOpdf
   \usepackage[pdftex]{graphicx}
   \graphicspath{{../pdf/}{../jpeg/}}
   \DeclareGraphicsExtensions{.pdf,.jpeg,.png}
\else
  \usepackage[dvips]{graphicx}
  \graphicspath{{../eps/}}
  \DeclareGraphicsExtensions{.eps}
\fi
\usepackage{amssymb}

\newcommand*\rotten{\rotatebox{2}}
\newcommand*\rot{\rotatebox{90}}

\usepackage{amsmath}
\usepackage{array}
\begin{document}
%
\title{\thename: Real-time Multi-Person 3D Pose Estimation from Multiple Views}

\author{\IEEEauthorblockN{Alessio Elmi \textcircled{r} Davide Mazzini \textcircled{r} Pietro Tortella\thanks{\textcircled{r} Equal contribution. Authors order determined by random function.}}
\IEEEauthorblockA{Checkout Technologies s.r.l.\\
20100 Milan, Italy\\
Email: \{alessio, davide, pietro\}@checkoutfree.it}}


%


\maketitle

\begin{abstract}
We present an approach to perform 3D pose estimation of multiple people from a few calibrated camera views. Our architecture, leveraging the recently proposed unprojection layer, aggregates feature-maps from a 2D pose estimator backbone into a comprehensive representation of the 3D scene. Such intermediate representation is then elaborated by a fully-convolutional volumetric network and a decoding stage to extract 3D skeletons with sub-voxel accuracy. Our method achieves state of the art MPJPE on the CMU Panoptic dataset using a few \emph{unseen} views and obtains competitive results even with a single input view. We also assess the transfer learning capabilities of the model by testing it against the publicly available Shelf dataset obtaining good performance metrics. The proposed method is inherently efficient: as a pure bottom-up approach, it is computationally independent of the number of people in the scene. Furthermore, even though the computational burden of the 2D part scales linearly with the number of input views, the overall architecture is able to exploit a very lightweight 2D backbone which is orders of magnitude faster than the volumetric counterpart, resulting in fast inference time. The system can run at 6 FPS, processing up to 10 camera views on a single 1080Ti GPU. 
\end{abstract}


%
\IEEEpeerreviewmaketitle

\section{Introduction}

Multi-person 3D pose estimation is a complex problem, with many applications in different fields of computer vision, like people tracking or augmented reality. This problem is usually tackled with a two steps approach. At first, every view is processed independently in order to produce a set of 2D poses - or possibly, some intermediate feature representation. In this stage, all the achievements in 2D pose estimation field can be exploited (see \cite{dang2019deep} for a survey). Next, these poses have to be matched across views and eventually triangulated, in order to produce a final estimate of the 3D scene. Usually, occlusions between people - or even self-occlusions - are the main difficulties to deal with: crowded scenes and complex poses produce noisy 2D detections, which are hard to filter out or recover in the matching and triangulation phase.

Hence, the idea of creating a system which is able to handle occlusions in a global way, and that is not affected by the limitations brought by single-view inferences. Inspired by \cite{iskakov2019learnable}, \cite{joo2015panoptic} and \cite{cao2017realtime}, we developed a multi-person 3D reconstruction system, which takes a set of images capturing the scene from different views and outputs a set of 3D pose reconstructions in a global reference frame. Its main building block is a fully convolutional neural network, where low-level features of the input views are unprojected, fused and transformed in order to produce a 3D representation of the probabilistic space. Following the general bottom-up approach of pose estimation, we extended the notion of \textit{part affinity field} in three dimensions, making the pose reconstruction from density maps quick and agile. By doing this, we avoided all the limitations of the top-down strategies, where scalability is penalized as the number of people grows, and where inter-person occlusions and self-occlusions cannot be encoded - and recovered - by the network in a global way. On the contrary, thanks to the huge variety of pose configurations available in the CMU Panoptic dataset and with clever augmentation strategies about view-points, we could prove that our system is not affected by those limitations: our system can exploit \textit{activations} and ``shadows" in the feature space to estimate occlusions. Moreover, it does not depend on sophisticated algorithms of \textit{detection-view assignment}, and it does not pay the computational burden of adding more views and subjects in the reconstruction process. Furthermore, we found that our system can produce good results even with just a single view suggesting that this approach can be further investigated also for monocular depth estimation tasks with multiple poses.

We conducted several experiments, which show the feasibility of our work and compare it to the other state-of-the-art approaches.

\begin{figure}[!t]
\includegraphics[width=\columnwidth]{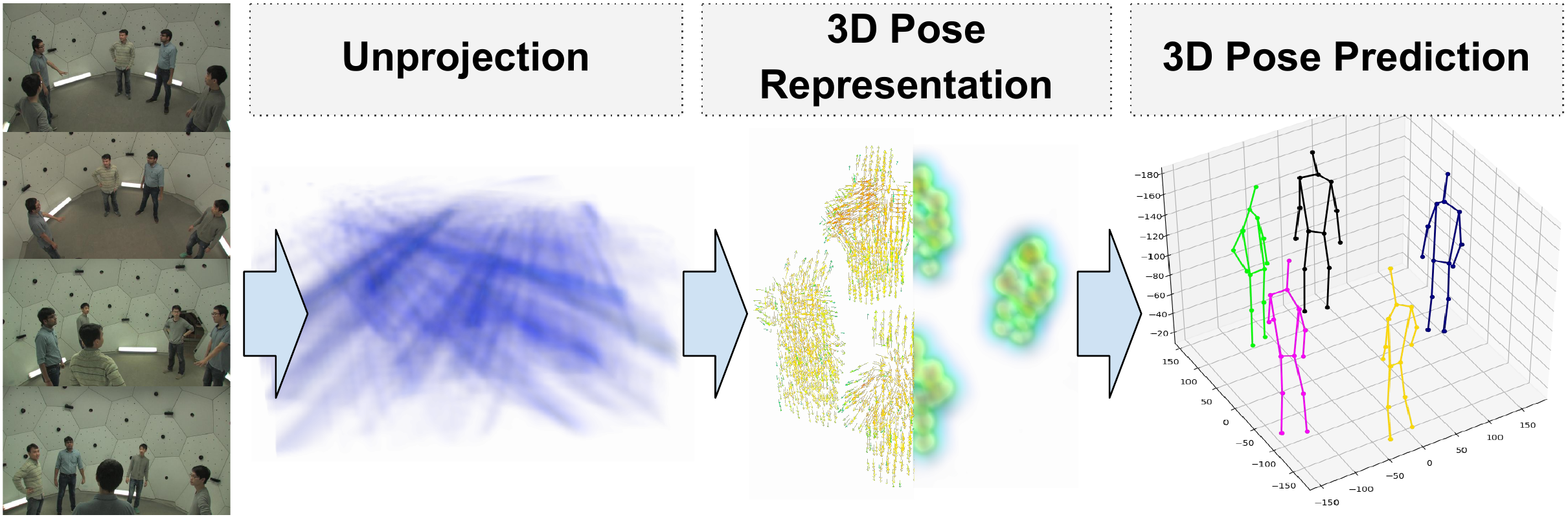}
\caption{This work proposes a fast and scalable approach for multi-person 3D pose estimation. Feature representations extracted from each views are aggregated and exploited to perform unified triangulation and pose estimation.}
\label{fig:teaser}
\end{figure}

\begin{figure*}[!t]
\includegraphics[width=\textwidth]{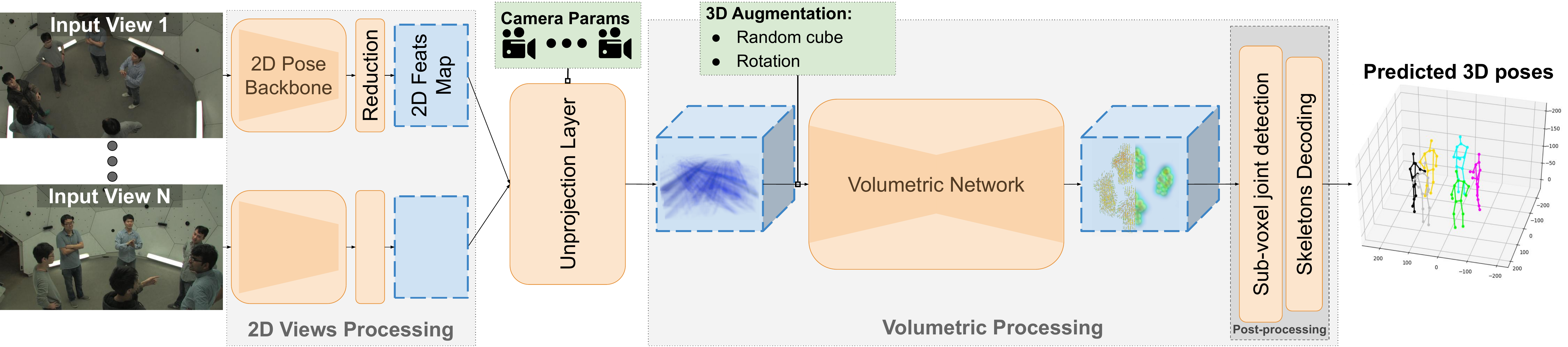}
\caption{An overall view of the complete processing pipeline. 2D pose backbone replicas process each view separately. Feature maps are then aggragated by the Unprojection layer into a 3D input representation of the scene. A volumetric network produces an output representation. A further decoding produces the final 3D pose estimations.}
\label{fig:architecture}
\end{figure*}

Our main contributions are the following:
\begin{itemize}
\item as far as we know this is the first complete bottom-up approach adopted in this context. In particular, it is capable of handling crowded scenes with good accuracy results and computational time.
\item We show that even a very light backbone can produce good results. This implies that adding more views is almost computationally free.
\item We introduce 3D-data augmentation policies that greatly enhance the number of samples seen by the volumetric network.
\item Our post-processing strategy leads to a sub-voxel localization, overcoming the issue of a quantized 3D space.
\end{itemize}

\section{Related Works}
Multi-view, multi-person 3D pose estimation tries to fuse the achievements coming from 2D pose estimation, structure from motion and monocular depth estimation research fields. All of them are very well studied and pretty active topics nowadays.

Pose estimation from a single image is usually tackled following one of these two main strategies: \textit{bottom-up} or \textit{top-down} approaches. The former try to infer all key-points (i.e. \textit{parts}) and/or limbs simultaneously and aggregate them eventually, using specific post-process logic. These methods can claim higher speed over their competitors since neural inference is done only once. At the same time, they usually have to deal with down-scaled feature maps, which limit the accuracy in terms of localization. In this group, we cannot omit \cite{cao2017realtime}. Inspired by the work of  \cite{wei2016convolutional}, they introduced the notion of part-affinity fields. Their work has been extended by \cite{newell2017associative, papandreou2018personlab} leading to a better part association, and by \cite{kreiss2019pifpaf}, where stronger descriptors led to a finer sub-pixel resolution. Other insights on the resolution issue were provided by \cite{feng2019}, with heatmap encoding/decoding refinements. On the contrary, top-down approaches \cite{newell2016stacked, chen2018cascaded, xiao2018simple}, possibly combined with multi-scale strategies \cite{ke2018multi, sun2019deep}, rely on object detectors to identify humans in the scene, then a single-person neural inference is performed for each of them. These techniques generally outperform their bottom-up competitors on public challenges, while suffering in scalability with increasing number of subjects. Some hybrid approaches have emerged as well \cite{kocabas2018multiposenet}. Finally, we want to mention some attempts \cite{osokin2018lightweight_openpose, zhang2019cvpr} to reduce the computational burden of pose estimation networks.

Three-dimensional pose estimation has emerged following two different tracks. The first one aims to recover the third dimension from a monocular view \cite{tome2017lifting, pavllo20193d, yang20183d, sun2018integral, nibali20193d, kanazawa2019learning, habibie2019wild}. These methods usually start from 2D pose estimations, and lift them in a second stage in order to obtain their depth. In particular, they all deal with single-pose scenarios. We mention two attempts to extend this task to a multiple poses: Moon et al. \cite{moon2019camera} adopted a top-down strategy; Rogez et al. \cite{rogez2019lcr} introduced pose proposals (from anchor-poses) in the spirit of the Faster R-CNN approach. The second research track takes advantage of multiple views and claims to reconstruct 3D poses in a global reference frame. Sometimes this is the initial step of detection-to-track pipelines, like in \cite{bridgeman2019multi, ohashi2020synergetic}, where temporal evolution can be exploited in order to refine predictions. Multiple-view pose reconstruction  may focus on single \cite{martinez2017simple, pavlakos2017harvesting, tome2018rethinking, qiu2019cross, iskakov2019learnable} or multiple poses \cite{pirinen2019domes}, and they can exploit geometrical constraints \cite{belagiannis20143d, tanke2019iterative}, in pair with visual features \cite{dong2019fast}. In particular, we highlight two works where multi-view projections have been combined with deep learning. \cite{qiu2019cross} exploited the epipolar geometry in order to refine 2D pose estimation model, and consequently improve the final single pose 3D reconstruction. \cite{iskakov2019learnable}  showed that 2D features of each view can be fused and processed into a volumetric representation, which is analyzed to achieve a neat 3D reconstruction of the pose - again - for a single subject. However, to the best of our knowledge there is not any attempt to extend this approach to a multi-person scenario.

\section{Method}\label{sec:method}

We call \thename\ our system.
In this section, we outline the architecture of \thename,
followed by a detailed explanation of all its components.

We are given a \emph{detection space} $S$ with fixed boundaries
and a set of fixed \emph{setup cameras}
$\{C_i\}_{i = 1, \ldots, N_c}$ whose intrinsic and extrinsic parameters
are known. In particular, the projections $P_i: S \to F_i$ are known,
where $F_i$ denotes the frame of the camera $C_i$.
The cameras are synchronized, so for any time $t$ we have a set of images
$I^t_1,\ldots, I^t_{N_c}$, one for each camera. We will assume the time
fixed throughout the paper, and omit the superscript $t$.

The input of \thename\  is a set of pairs $\{I_i, C_{\nu_i}\}_{i=1,\ldots,m}$
where $I_i$ is an image and $C_{\nu_i}$ is one of the setup cameras.
The number of input pairs $m$ is variable and can range from 1 to $N_c$.
In Section \ref{sec:ablation} we study both from the performance and 
computational sides the impact of the number of input views.

The output of \thename\  is a set of 3D human poses $\{A_1, \ldots, A_k\}$,
with $k$ an arbitrary number.
A 3D human pose $A_i$ is a list $(a_i^l)_{l \in pose\_layout}$ of \emph{joints}.
Each joint is a pair composed of a point in the space $S$ and a label identifying
the joint type, but when no confusion arises we identify the joint with
the underlying point in $S$.
The joint type ranges in a \emph{pose layout} named CMU14, described in Section \ref{sec:implementation}.

The internal pipeline of \thename\  is composed of three main stages (see Figure \ref{fig:architecture}):

\begin{itemize}
    \item a \emph{2D Views Processing} stage which returns a 2D feature map for each camera;
    \item an \emph{Unprojection layer} \cite{iskakov2019learnable} which aggregates the information coming from all the 2D views into a 3D features space representation;
    \item a \emph{Volumetric Processing} that process the aggregated 3D representation and produces the output;
\end{itemize}
and each of these stages are composed of a different number of modules.

\subsection{2D Views Processing}

This processing stage takes as input one image $I$
and produces a 2D activation $\mathcal{R}(\mathcal{B}(I))$.
When \thename\  processes a set
of pairs $\{(I_i, C_i)\}$, each $I_i$ is fed independently to the 2D Views 
processing stage. The different 2D View Processing stages share the same
weight.

The stage is composed of two modules: a \emph{2D Pose Backbone} followed by a 
\emph{Reduction module}. 

\subsubsection{2D Backbone}

The input to the 2D Backbone module is an image $I$, and the
output is a 2D feature map $\mathcal{B}(I)$. The 2D backbone is a MobileNet V1 \cite{howard2017mobilenets} with some modifications from \cite{osokin2018lightweight_openpose} on the latest layers. The stride of \emph{conv4\_2\/dw} has been removed and all succeeding convolutions have been set to dilation 2. This operation makes the network global stride to be $16$ instead of $32$ which is common for classification networks. We used weights pretrained on COCO dataset from \cite{osokin2018lightweight_openpose}.

\subsubsection{Reduction Module} 

Input to the reduction module is the 2D feature map $\mathcal{B}(I)$, and
the output is a 2D feature map $\mathcal{R}(\mathcal{B}(I)$.
The purpose of this module is to project the feature space produced 
by the 2D Backbone to a lower-dimensional feature 
space. This module is crucial in order to encode the information
of the backbone into a lighter feature map, to maintain the computations
performed by the Volumetric Network feasible. 
Our Reduction Module is essentially a residual module composed of three 
depth-wise convolutions + ReLUs.
We borrow this 
architecture from \cite{osokin2018lightweight_openpose}.

\subsection{Unprojection}

This processing stage represents the contact point between the 2D feature 
maps and the 3D model of the scene, collecting the result of the 
2D Processing stage into a 3D feature map representation.
This is the only stage of \thename\ 
that uses the calibration parameters of the cameras, and it has no
trainable parameters.

Fix integer numbers $Q_x, Q_y, Q_z$, and a positive float value $Q_{size}$.
Construct a cube $\mathrm{C} \subseteq S$ composed of 
$Q_x \times Q_y \times Q_z$
voxels with edge of length $Q_{size}$. In $\mathrm{C}$ one has the integer 
coordinate system $(i_x, i_y, i_z)$ corresponding to the index of the voxels
of $\mathrm{C}$, and we denote by $\iota: \mathrm{C} \to S$
the embedding.

The input to this stage is a set of pairs
$\{(\mathcal{R}_i, C_{\nu_i})\}_{i=1,\ldots,m}$, where each $\mathcal{R}_i$
is the output
of one of the 2D View Processing modules, and $C_{\nu_i}$ is one of the
setup cameras.

The output of the unprojection stage is a 3D feature map $\mathcal{U}^\iota$ 
with shape $Q_x \times Q_y \times Q_z \times N_{feats}$, where
$N_{feats}$ is the number of channels of the 2D feature map $\mathcal{R}$.

To compute the value of the $j$-th feature of the voxel 
$\mathcal{U}^\iota(i_x,i_y,i_z)$
we use the formula:
\begin{equation}
    \U^\iota(i_x, i_y, i_z)^j = 
    \frac{1}{m} 
    \cdot \sum_{i=1}^m \R_i(P_{\nu_i}(\iota(i_x, i_y, i_z)))^j
\end{equation}
where recall that $P_i$ denotes the projection associated with
the camera $C_i$, and by $\mathcal{R}_i(u, v)^j$ we denote the $j$-th channel
of the 2D feature map $\mathcal{R}_i$ at the point with frame coordinates $(u, v)$.

This layer is a generalization of the Unprojection introduced in
\cite{iskakov2019learnable} where a cube is built around each person.
It can be efficiently implemented using 
vectorized operations and a differentiable sampling operator \cite{jaderberg2015spatial}.

\subsection{Volumetric Processing}

Input to this stage is the 3D feature map $\mathcal{U}$ output of the Unprojection layer.

The output of this stage is a set of 3D human poses 
$\{A_1,\ldots,A_k\}$.

The stage is composed of three modules: 
\begin{itemize}
    \item the \emph{Volumetric Network},
    \item the \emph{Sub-voxel Joint Detection}
    \item the \emph{Skeleton Decoder}.
\end{itemize}
The approach is similar to
OpenPose \cite{cao2017realtime}: the neural part of the network is 
trained to predict a Gaussian centered on each joint; the network 
should also predict a set of Part Affinity Fields (PAFs) that are 
used by the decoder to efficiently build the skeletons. 
Our method directly predicts 3D poses, thus the main differences 
between our volumetric processing part and OpenPose are in the use 
of a different neural 
architecture to handle 3D volumes data, and 
an adaptation to the 3D setting of the decoding of the output
of the Volumetric network.
Moreover, we introduce a Sub-voxel Peak Detector module
to increase the accuracy of the joints predictions.

\subsubsection{Volumetric Network}\label{sec:volumetric_network}

This is the trainable neural part of the volumetric processing. 
The purpose is to predict a set of 3D Gaussians centered on every joint 
and a set of 3D PAFs for the skeleton reconstruction.

The input to this module is the 3D activation $\mathcal{U}$ output of
the unprojection layer with shape 
$Q_x \times Q_y \times Q_z \times N_{feats}$.

Output of this module is a 3D activation $\mathcal{V}$ with shape
$Q_x \times Q_y \times Q_z \times N_{gt}$, where 
$N_{gt} = N_{joints} + 3 \cdot N_{PAF}$, where 
$N_{joints}$ is the number of joints of the pose layout, and 
$N_{PAF}$ is the number of PAF.
This output can also be seen as a pair of collections
$\V = ((H^l)_{l \in pose\_layout}, (\mathbf{V}^s)_{s \in PAFs})$
where each $H^l$ is a 3D feature map corresponding to a heatmap
and each $\mathbf{V}^s$ is a collection of 3 (one per each of to the
3-dimensional directions of the vector)
3D features map corresponding 
to a vectormap.

We adopt a V2V network from \cite{moon2018v2v}, but we set
the minimum number of channels of the earliest and latest 
layers to $64$ in wherever layer the original network has $32$ channels. 
We name this modified V2V network: \emph{V2V64}. We also
experimented with 32 and 96 channels architectures.
Results are reported in Section \ref{sec:ablation}.

The output $\V$ of the module is then confronted with the ground-truth
with an appropriate loss function, which is used to perform the training
of \thename.
The dataset labels are lists of poses of persons in the 3D space. 
The procedure to create ground-truth heatmaps and vectormaps 
is a generalization to 3D space of the one in \cite{cao2017realtime},
so we omit the details.
We opted to use a SmoothL1 loss function and to weight 
equally the loss coming from the heatmap and the vectormap. 
We experimented different loss functions and weights between heatmap
and vectormap, the results are reported in \ref{sec:ablation}.

\subsubsection{Sub-voxel Joint Detector}
Several state-of-the-art works on single-person pose estimation  are based on 
a variation of the Integral Regression Framework 
\cite{sun2018integral, iskakov2019learnable, sun2019deep}
which represents the unifying approach between heatmap 
and regression-based methods.
The Integral Pose Regression framework assumes that 
the point to be localized follows a unimodal distribution.
This is not the case 
of multiple poses scenarios, where more than one peak need 
to be estimated.
We present an alternative formulation of such
framework which, under the correct assumptions, can be used 
in a multi-person setup.

The sub-voxel joint detector module takes as input one
heatmap $H$ ouput of the Volumetric network, and outputs
a list of peaks $\mathcal{S}(H) = \{p_i\}$. The module is applied to
each joint heatmap $\{H^l\}$, obtaining a set of peak
for each joint type $\{\{p_i^l\}_{i=1,\ldots,n_l}\}_l$.

In order to simplify the notation, we discuss the 1D case, 
but operators can be intuitively extended to 2D or 3D. 
Given a learned heatmap $H$,
for each spatial location $x$ the values $H(x)$ 
represent the probability of such location of being a joint.
We fix a neighbour function $N: \mathrm{C} \to subsets(\mathrm{C})$
that associates to each point a neighbor of it
(typically, an interval of a given radius centered at $x$).
Define the non-local maxima suppression 
$\mathbf{P}: \mathrm{C} \to \{0, 1\}$ via the formula :
\begin{equation}
    \textbf{P}(x) = 
    \delta \left(
        \left(
        \max_{ \bar x \in N(x)} \textbf{H}( \bar x) 
        \right) 
        = \textbf{H}(x)
    \right)
\end{equation}
where $\delta$ is a Dirac function. $\mathbf{P}(x) = 1$ if and only if
$x$ is a maximum of $H_{|N(x)}$.
Define the \emph{pixel-peaks} as 
$$
\tilde{\mathrm{R}} \ =\  \{ x \in \mathrm{C} \ | \ P(x) = 1\}
$$
For each $x \in \hat{\mathrm{R}}$, define the localized heatmap
$$
L_xH = \frac{1}{\sum_{\bar{x} \in N(x)} H(\bar{x})} \cdot H_{|N(x)}
$$
Finally, define the \emph{sub-pixel peaks} as 
$$
\mathcal{S}(H) = \left\{ 
\ \sum_{ \bar{x} \in N(x)}{ \bar{x} \cdot (L_xH) (\bar{x})} \ 
| \ x \in \tilde{\mathrm{R}}  
\right\}
$$
The assumption we rely on is that for every $x$, in the
neighbour $N(x)$ there should be at most one local maximum.
In general, this assumption holds if the radius is small enough 
w.r.t. the quantization constant $Q_{size}$. 
In practice, we obtain good results by choosing
$N(x)$ to be a 1 or 2 voxels radius interval centered at $x$,
see Section \ref{sec:ablation}.

\subsubsection{Skeletons decoder} 
This module takes as input the peaks 
$\{\mathcal{S}(H^l)\}_{l \in pose\_ layout}$ of the sub-pixel joint
detection and the vectormaps $\{\mathbf{V}^s\}_{s \in PAFs}$ 
output of the Volumetric Network and
outputs a list of 3D poses. 
Our algorithm is a direct extension of the one proposed by OpenPose
\cite{cao2017realtime}, with the only difference that line integrals 
are computed over three-dimensional vector fields.

\begin{table}[!t]
\renewcommand{\arraystretch}{1}
\setlength{\tabcolsep}{3pt}
\caption{Ablation studies on PanopticD2D validation set for different aspects of our architecture. 3D Augmentations, Loss type and ratio between heatmap loss and vectormap loss weights.}
\label{tab:ablation}
\centering
\resizebox{1\columnwidth}{!}{
\begin{tabular}{c c|c|c c c c c c|c}

& & & \multicolumn{6}{c}{PCP} & \\
& & $\begin{matrix} \text{MPJPE} \\ \text{(cm)} \end{matrix}$ & \rot{Head} & \rot{Torso} & \rot{Up Arm} & \rot{Lo Arm} & \rot{Up Leg} & \rot{Lo Leg} & \rot{Avg}\\
\hline
\multicolumn{10}{c}{\vspace{0.1cm}}\\
Cube & \multicolumn{1}{c}{\rotten{Rotation}} & \multicolumn{8}{c}{3D Augmentations}\\
\hline
            & & 8.236 & 99.1 & 99.3 & 87.8 & 65.4 & 96.9 & 88.3 & 89.2 \\
 \checkmark & & 4.598 & 99.6 & \textbf{99.7} & 98.5 & 90.1 & \textbf{99.3} & 98.5 & 97.7 \\
 & \checkmark & 5.350 & 99.6 & \textbf{99.7} & 98.6 & 91.1 & 99.0 & 94.9 & 97.3 \\
 \checkmark & \checkmark & \textbf{3.859} & \textbf{99.7} & \textbf{99.7} & \textbf{99.5} & \textbf{95.6} & \textbf{99.3} & \textbf{98.8} & \textbf{98.8}\\
\hline
\multicolumn{10}{c}{\vspace{0.1cm}}\\
\multicolumn{2}{c}{} & \multicolumn{8}{c}{Number of Volumetric Features}\\
\hline
\multicolumn{2}{c|}{32} & 4.760 & 99.6 & 99.7 & 97.1 & 78.9 & 99.5 & 98.6 & 95.9\\
\multicolumn{2}{c|}{64} & \textbf{3.859} & \textbf{99.7} & \textbf{99.7} & \textbf{99.5} & 95.6 & \textbf{99.3} & \textbf{98.8} & 98.8\\
\multicolumn{2}{c|}{96} & 3.975 & \textbf{99.7} & \textbf{99.7} & \textbf{99.5} & \textbf{96.2} & \textbf{99.3} & 98.7 & \textbf{98.9}\\
 \hline
\multicolumn{10}{c}{\vspace{0.1cm}}\\
\multicolumn{2}{c}{} & \multicolumn{8}{c}{Loss Type}\\
\hline
\multicolumn{2}{c|}{L1} & 4.106 & 99.6 & \textbf{99.7} & 99.2 & 96.2 & 99.0 & 98.0 & 98.7\\
\multicolumn{2}{c|}{L2} & 4.125 & 99.6 & \textbf{99.7} & \textbf{99.5} & \textbf{96.6} & \textbf{99.4} & \textbf{98.9} & \textbf{99.0}\\
\multicolumn{2}{c|}{SmoothL1} &  \textbf{3.859} & \textbf{99.7} & \textbf{99.7} & \textbf{99.5} & 95.6 & 99.3 & 98.8 & 98.8\\
 \hline
 \multicolumn{10}{c}{\vspace{0.1cm}}\\
\multicolumn{2}{c}{} & \multicolumn{8}{c}{Heatmap / Vectormap Loss Ratio}\\
\hline
\multicolumn{2}{c|}{1} &  \textbf{3.859} & \textbf{99.7} & \textbf{99.7} & \textbf{99.5} & 95.6 & 99.3 & \textbf{98.8} & 98.8\\
\multicolumn{2}{c|}{3} & 4.074 & \textbf{99.7} & \textbf{99.7} & 99.1 & \textbf{96.6} & \textbf{99.5} & 98.6 & \textbf{98.9}\\
\multicolumn{2}{c|}{10} & 3.935 & \textbf{99.7} & \textbf{99.7} & 98.0 & 90.9 & \textbf{99.5} & \textbf{98.8} & 97.9\\
 \hline
 \multicolumn{10}{c}{\vspace{0.1cm}}\\
\multicolumn{2}{c}{} & \multicolumn{8}{c}{Sub-voxel refinement}\\
\hline
\multicolumn{2}{c}{} & 4.899 & \textbf{99.7} & \textbf{99.7} & 99.4 & 94.9 & \textbf{99.3} & \textbf{98.8} & 98.6 \\
\multicolumn{2}{c}{\checkmark} & \textbf{3.859} & \textbf{99.7} & \textbf{99.7} & \textbf{99.5} & \textbf{95.6} & \textbf{99.3} & \textbf{98.8} & \textbf{98.8}\\
 \hline
\end{tabular}
}
\end{table}

\begin{figure}[!t]
\includegraphics[width=\columnwidth]{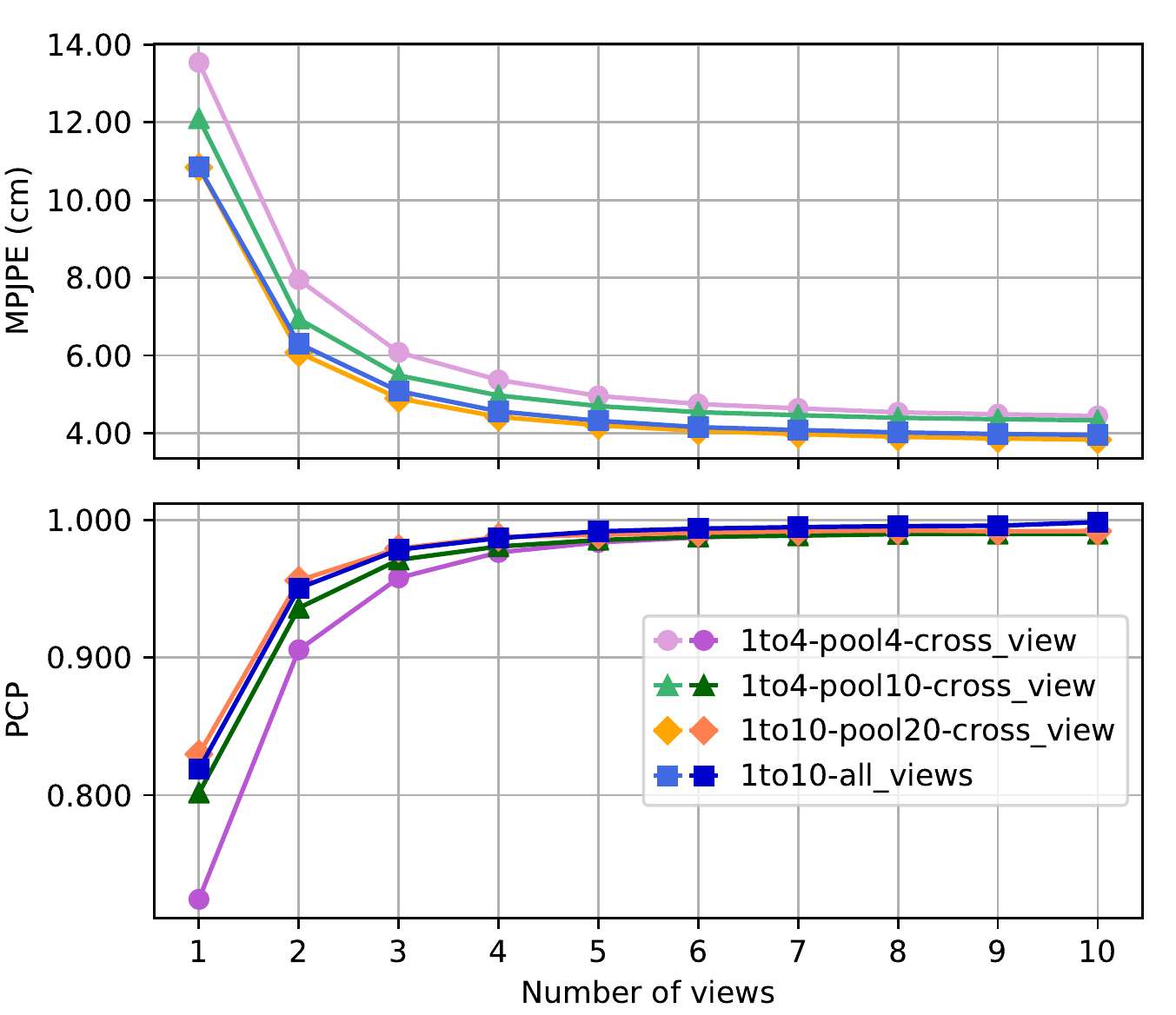}
\caption{Accuracy vs number of views. \textit{1to4-pool4-cross\_view}: training with 1 to 4 simultaneous views from a pool of 4; \textit{1to4-pool10-cross\_view}: 1 to 4 from a pool of 10; \textit{1to10-pool20-cross\_view}: 1 to 10 from a pool of 20; \textit{1to10-all\_views}: 1 to 10 from all available views. Configurations 1-2-3 are \textit{cross-view}.}
\label{fig:accuracy_vs_views}
\end{figure}

\section{Experimental setup}
\subsection{Datasets}

\subsubsection{CMU Panoptic dataset \cite{joo2015panoptic}} it consists of 31 Full-HD and 480 VGA video streams from synchronized cameras at 29.97 FPS; various scenes (65 sequences with multiple people, social interactions, and a wide range of actions) for a total duration of 5.5 hours. The dataset includes robustly labeled 3D poses, computed using all the camera views. This dataset is perhaps the most complete, open and free to use dataset available for the task of 3D pose estimation. However, considering that they released annotations quite recently, most works in literature use it only for qualitative evaluations \cite{dong2019fast} or for single-person pose detection \cite{iskakov2019learnable} discarding multi-person scenes. To the best of our knowledge only \cite{pirinen2019domes} makes use of CMU Panoptic dataset to train and evaluate multi-person 3D pose estimation. We adopt the same subset of scenes and the same train/val/test split of CMU Panoptic used in \cite{pirinen2019domes}: 20 scenes (343k images)  of which 10, 4 and 6 scenes for training, validation and test respectively. Only HD cameras are used with data frame rates downsampled to 2 FPS. Since one of our concerns is to assess the cross-view generalization of our model, we split the dataset by scene and by view. Val and test splits use cameras 2, 13, 16, 18 while the train split uses all (or a subset of) the remaining 27 cameras. This is the same camera split used by \cite{iskakov2019learnable}. We name this dataset: \textit{PanopticD2D}.

\subsubsection{Shelf \cite{belagiannis20143d}} we adopt this dataset to evaluate the ability of our model to transfer to a completely unseen setup. It consists of a single scene of four people disassembling a shelf at a close range. Video streams are from five calibrated cameras. The dataset includes 3D annotated groundtruth skeletons.
\subsection{Evaluation metrics}
We employed two commonly used metrics that capture different types of errors in models prediction:
\begin{itemize}
    \item \emph{MPJPE: Mean Per Joint Precision Error}. Given a pair of skeletons, MPJPE is defined as the average of the square distance of the predicted joints from the corresponding ground-truth joints.
    \item \emph{PCP: Percentage of Correct estimated Parts}. We implemented this metric according to \cite{dong2019fast}. A body part is correct if the average distance of the two joints is less than a threshold from the corresponding groundtruth joints locations. The threshold is computed as the 50\% of the length of the groundtruth body part.
\end{itemize}
Before computing these metrics we associate for each scene the predicted skeletons to the groundtruth skeletons using linear assignment.

\subsection{Implementation details}\label{sec:implementation}
\subsubsection{Pose Layout}
We used a simplified pose layout of 14 keypoints. Apart from the canonical 12 parts of arms and legs, we only added \textit{neck} and \textit{nose}. Sometimes, a layout conversion was needed across different datasets and labeling standards. Moreover, we defined 13 PAFs; starting from the \textit{neck}, a tree-structure along arms, legs and nose has been defined. In our setup, increasing excessively the number of joints or PAFs would not make sense due to the limitations of our quantized space.
\subsubsection{Skeleton Decoding}
Parameters have been found by performing a grid search on Panoptic D2D validation set. Eventually, we opted for an interpolation over a region of size $5 \times 5 \times 5$ voxels. Then, all local maxima with a score lower than 0.3 are discarded; every PAF where the linear integral is on average lower than 0.2 is also removed. Finally, only candidate poses with more than 7 keypoints are retained.
\subsubsection{3D space quantization} we set the size of the quantization voxel to 7.5 cm. This allows us to maintain a quantization of $64 \times 64 \times 32$ voxels on Panoptic dataset to efficiently cover the whole scene of approximately $5 \times 5 \times 2.5$ meters, the last dimension being the vertical axis.
\subsubsection{Training recipe}
Models have been trained with Adam optimizer. We set the initial learning rate to 0.002 and used a step decay policy of 0.3 every 50 epochs. All models have been trained for 200 epochs with a batch size of 8. We implemented the architecture in PyTorch.
\section{Ablation analysis}\label{sec:ablation}

\subsection{3D Augmentation}
We applied 3D data augmentation techniques to the 3D feature space between the Unprojection layer and the Volumetric network. In particular, we implemented the followings:

\subsubsection{Random cube embedding} 
During the training, we consider $\mathrm{C} \subseteq S$ to be 
strictly smaller, and to be randomly embedded. This corresponds
to take a random crop of the 3D crop of the scene to
be considered for the parameters update.

From the volumetric network point of view, this reflects into a data 
augmentation strategy, since moving the cube inside $S$ corresponds 
to a change of the observed scene and a change 
in the extrinsic parameters of the cameras.

We set $\mathrm{C}$ to have $32 \times 32 \times 32$ voxels,
and we change the embedding at the start of each epoch.

\subsubsection{Random rotation} we implement rotations along the vertical axis of $90^{\circ}, 180^{\circ}, 270^{\circ}$, to allow a fast implementation. One should take care of the fact that rotation of the 3D space is not reflected into images transformation, so when a rotation is applied we cut the back-propagation graph just before the unprojection layer. In our specific architecture, this sparse back-prop signal does not drastically affect the training since the only trainable part before the volumetric network is the Reduction layer which has a limited number of parameters.

\begin{table}[!t]
\renewcommand{\arraystretch}{1}
\setlength{\tabcolsep}{3pt}
\caption{Methods comparison on Panoptic D2D test set. MPJPE and PCP metrics for scenes with single person and multiple people.}
\label{tab:d2d}
\centering
\resizebox{1\columnwidth}{!}{
\begin{tabular}{l|r r r|r}
\hline
 & \multicolumn{3}{c|}{MPJPE (cm)} & PCP \\
 Model & single & multi & avg & avg \\
 \hline
 ACTOR \cite{pirinen2019domes} (2 views)* & 17.21 & 50.24 & 33.72 & - \\
 ACTOR (4 views)* & 8.19 & 20.10 & 14.14 & - \\
 ACTOR (10 views)* & 6.13 & 12.21 & 9.17 & - \\
 Oracle \cite{pirinen2019domes} (using GT to select cameras)* & \textbf{4.24} & 9.19 & 6.71 & - \\
 \hline
 Ours (1 unseen view) & 10.34 & 9.32 & 9.43 & 80.8 \\
 Ours (2 to 4 unseen views depending on scene) & 5.30 & \textbf{4.09} & \textbf{4.22} & 98.2\\
 \hline
 Ours (10 views, from training view pool) & \textbf{3.50} & \textbf{3.56} & \textbf{3.55} & \textbf{98.6} \\
 \hline 
  \multicolumn{5}{c}{} \\
 \multicolumn{5}{p{9cm}}{*ACTOR: number in brackets refers to maximum number of views to choose from. Oracle means: best views to triangulate are selected using groundtruth.}
 \end{tabular}
 }
 \end{table}
 \begin{table}[!t]
\renewcommand{\arraystretch}{1}
\setlength{\tabcolsep}{3pt}
\caption{Quantitative comparison on Shelf dataset. Metric is PCP.}
\label{tab:shelf}
\centering
\begin{tabular}{l|c c c|c|c}
\hline
 Model & Actor 1 & Actor 2 & Actor 3 & Avg & Speed(s) \\
 \hline
 Belagiannis et al. \cite{belagiannis20143d} & 66.1 & 65.0 & 83.2 & 71.4 & -\\
 Belagiannis et al. \cite{belagiannis2014multiple} & 75.0 & 67.0 & 86.0 & 76.0 & -\\
 Belagiannis et al. \cite{belagiannis20153d} & 75.3 & 69.7 & 87.6 & 77.5 & - \\
 Ershadi et al. \cite{ershadi2018multiple} & 93.3 & 75.9 & 94.8 & 88.0 & - \\
  Dong et al. \cite{dong2019fast} & \textbf{98.8} & \textbf{94.1} & \textbf{97.8} & \textbf{96.9} & .465 \\
 Ours & 94.3 & 78.4 & 96.8 & 89.8 & \textbf{.146}\\
 \hline
 \end{tabular}
 \end{table}

\begin{figure}[!t]
\includegraphics[width=\columnwidth]{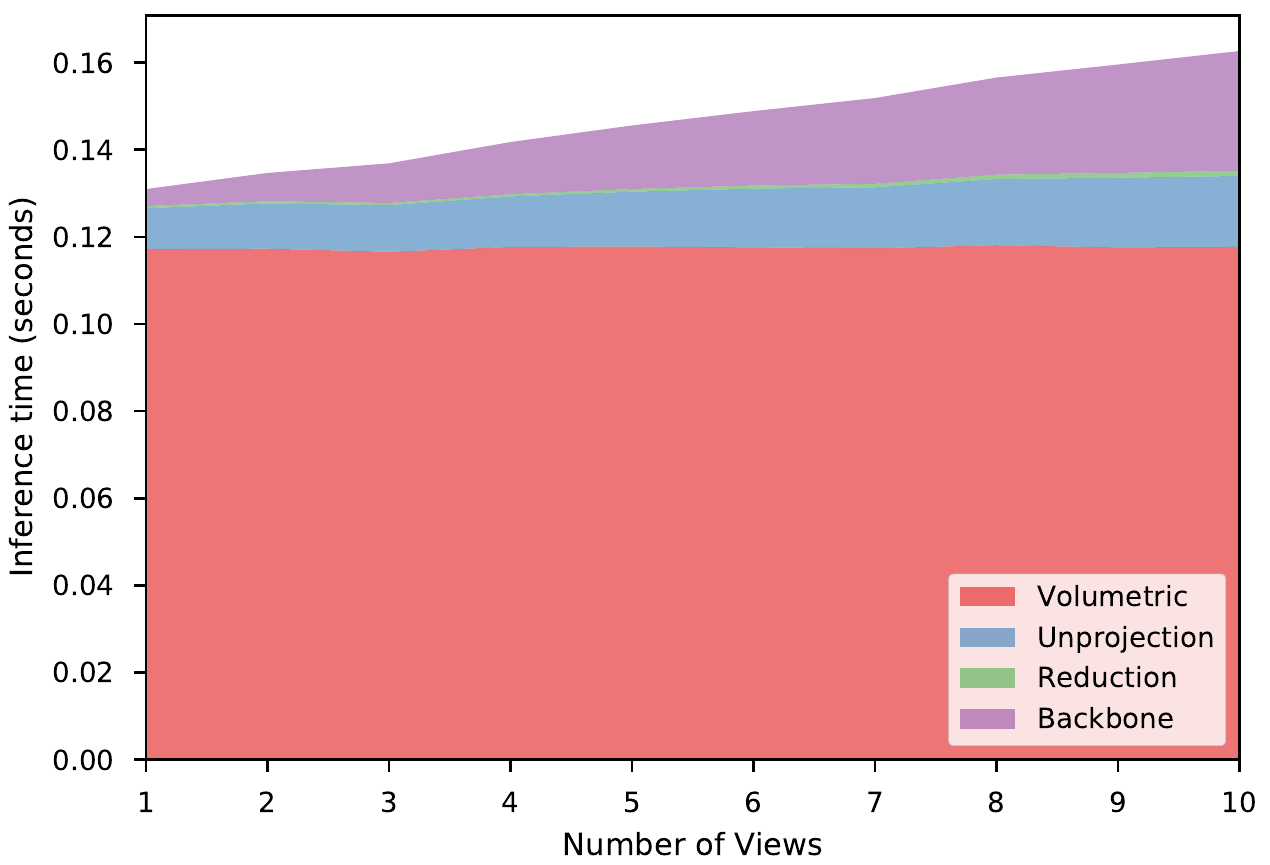}
\caption{Adding more views increases computational time by a linear factor. However, only few modules are affected by this growth. The main CNN block (in red) has a $O(1)$ complexity, both in the number of views and people. Inference time is measured on a single NVIDIA GeForce GTX 1080Ti.}
\label{fig:speed}
\end{figure}

\subsection{Architecture}
In Table \ref{tab:ablation} we reported the results of different experiments to evaluate the contribution of our architectural choices.

\subsubsection{Number of volumetric features} 
it refers to the channels of the volumetric input: it involves the 2D feature maps, the input/output of the unprojection  and the volumetric network. For 32 features we used the original V2V network whereas for 64 and 96 we modified it as described in Section \ref{sec:volumetric_network}. Models with 64 and 96 channels achieve similar MPJPE and PCP values but 64 is an obvious choice for being computationally lighter.

\subsubsection{Loss} 
we run experiments with different loss types and weighted differently the heatmap and vectormap losses. By evaluating separately PAFs and Peaks quality we noticed that good peaks have a stronger impact in the final metrics than good PAFs, thus we weighted more the Peak part of the loss. Results seem to suggest that the task of predicting good peaks should be tackled with a more elaborate approach than simply differentiate loss weights.

\subsubsection{Sub-voxel refinement} 
by activating it we achieve a lower MPJPE. It has almost no effect on PCP since it improves the sub-voxel localization but does not reduce false positives.

\subsection{Study on the number of input views} 
These experiments have a two-fold goal. On one side, we wanted to understand better the impact on the accuracy of a short/large number of views in the training pool; on the other hand, we wanted to check how well our augmentation strategies could compensate/emulate unseen angles. In Figure \ref{fig:accuracy_vs_views} we reported four experiments where we varied the number of views and the number of simultaneous angles used on each training inference. In particular, they show that even a few cameras can produce mildly good results; also, after a certain number, adding more views gives unnoticeable improvements.

\section{Comparison with state-of-the-art}
In Table \ref{tab:d2d} we report a comparison between our method and the results in \cite{pirinen2019domes} (ACTOR and ORACLE). We remark that the task that authors of \cite{pirinen2019domes} are trying to solve is different from ours. They train an agent to find what are the best views to use to triangulate \emph{that} particular scene. We consider it to be a good baseline even if the core task of \cite{pirinen2019domes} is not the triangulation algorithm itself.
We select 4 fixed validation views and we never train on those. Since some recordings have fewer views available, it turned out that only 36.2\% of the test set has 4 views, 31.3\% has 3 and 32.5\% has just two angles available.
The evaluation metric is the MPJPE expressed in $cm$. The MPJPE of our method is more than 3 times lower compared to ACTOR with 4 views and on average \emph{lower} than the Oracle. 

We also run our model on the Shelf dataset in order to test it in a completely new environment with unseen views, camera parameters, sensors and every other detail that can bias the evaluation. Results are reported in Table \ref{tab:shelf}. Our method obtains good results even if not on-par with the work by Dong et al. \cite{dong2019fast}. However, their approach is much slower being based on top-down 2D backbones. We detail a speed comparison between the two methods in Section \ref{sec:speed}.

\begin{figure}[!t]
\includegraphics[width=\columnwidth]{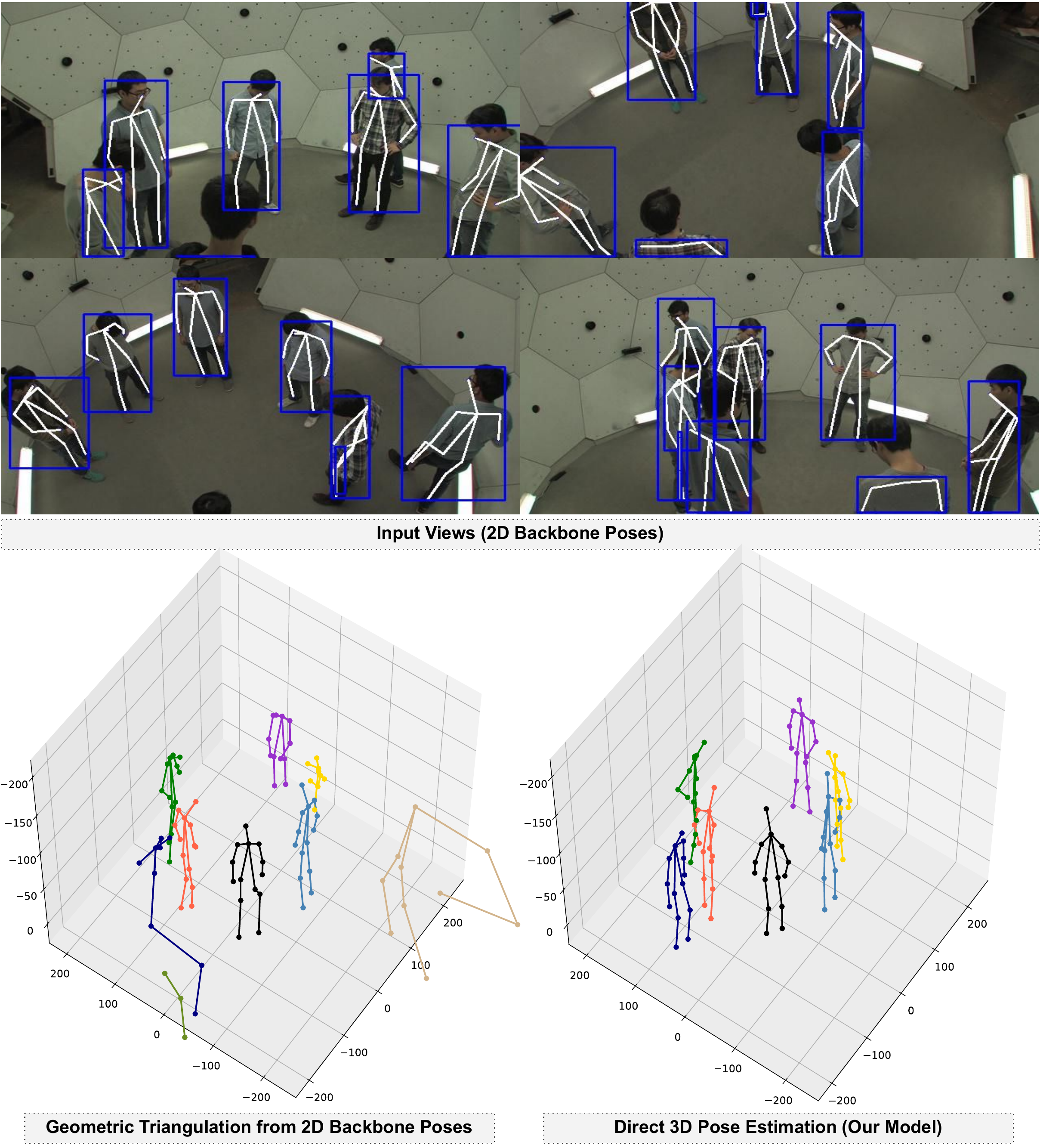}
\caption{Left: geometric triangulation using 2D poses from Lightweight OpenPose \cite{osokin2018lightweight_openpose} (same 2D backbone weights as ours) and iterative greedy matching \cite{tanke2019iterative}. Right: direct 3D pose estimation with our model.}
\label{fig:ours_vs_backbone}
\end{figure}
\begin{figure}[!t]
\centering
\includegraphics[width=\columnwidth]{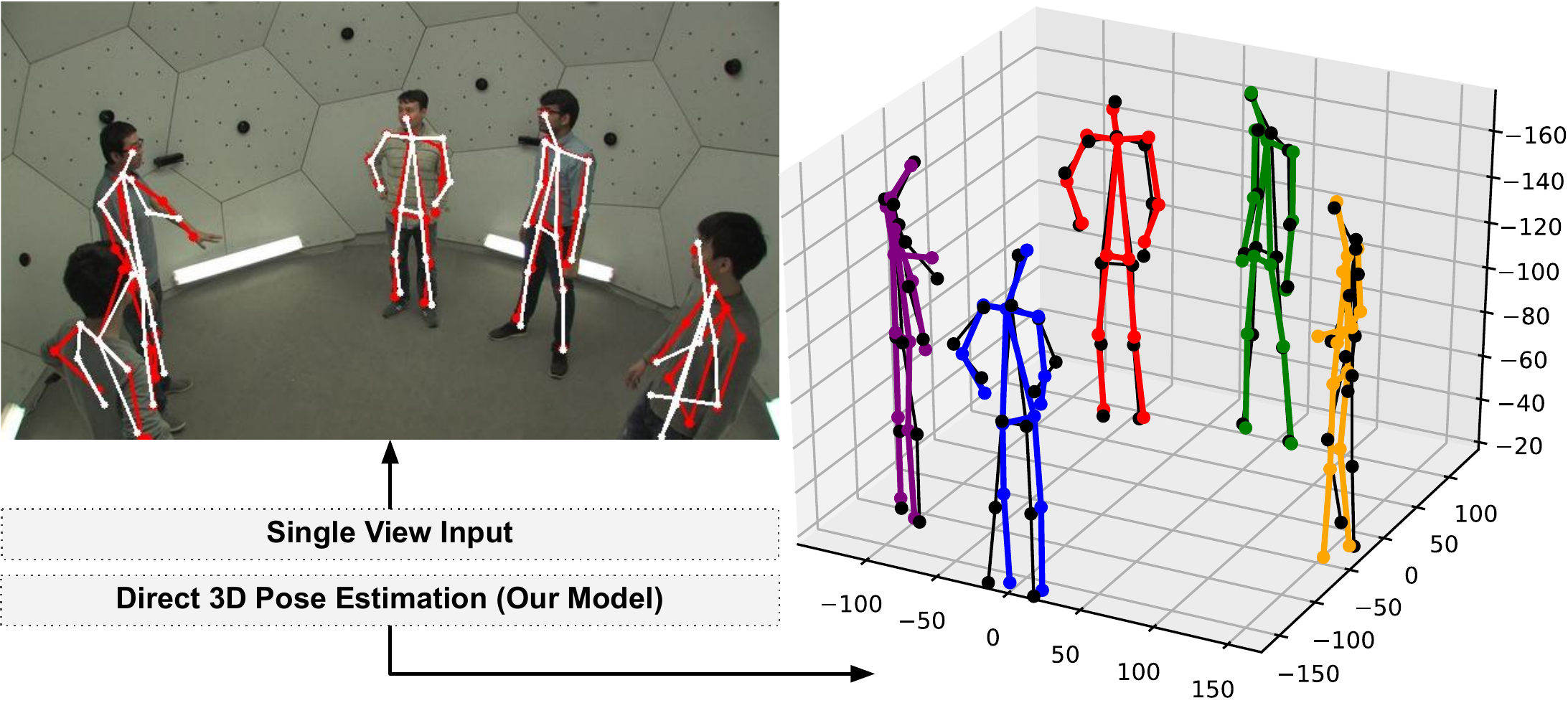}
\caption{Direct 3D pose predictions by our model from a single camera view. On the frame we projected in red the groundtruth, in white our predictions. In the 3D plot: predictions in color, groundtruth in dashed-black. The network ``hallucinates'' straight legs of non visible body parts relying on a strong learned prior.}
\label{fig:single_view}
\end{figure}

\subsection{Inference speed}\label{sec:speed}
Being a pure bottom-up approach, our method can scale well when increasing the number of views and subjects. Even though our complexity is $O(n)$ in the number of views and $O(n^2)$ in the number of people, adding more cameras affects only the \textit{Backbone}, \textit{Reduction} and \textit{Unprojection} modules, which are a small fraction of the cumulative computation burden (e.g. for 10 views they take all together only 45 ms, see Figure \ref{fig:speed}). On the other hand, post-processing the CNN output costs even less; Cao et al. \cite{cao2017realtime}, implemented an optimized version which takes 0.58 ms for a 9 people image. For reference we can compare our method with the one presented in \cite{dong2019fast}, see Table \ref{tab:shelf}. Their approach starts with a person detector \cite{li2017light}, which takes around 10 ms per view. Then, each detection is forwarded to two branches, of which the 2D pose estimation \cite{chen2018cascaded} is most expensive (we measured 67 ms). From here, the final 3D pose inference takes around 80 ms. We can estimate that a 5 views scenario with 5 people will take $(10 * 5) + (67 * 5) + 80 \approx 465 ms$, which is about ~3.2 times our implementation.

\subsection{Qualitative results}
In figure \ref{fig:ours_vs_backbone} we show a comparison between our model which performs direct 3D estimations and the result of the geometric triangulation using the 2D skeletons predicted by Lightweight OpenPose\cite{osokin2018lightweight_openpose} and the iterative greedy matching by \cite{tanke2019iterative}. Notice that our 2D backbone has exactly the same weights as the backbone of \cite{osokin2018lightweight_openpose} since we do not train nor finetune such part of the network. This highlights the power of estimating directly 3D poses: our volumetric architecture can learn strong pose priors and implicitly discards false detections. By exploiting the 3D representation of the space, it is less prone to occlusion-related errors and it can better deal with crowded scenes. This behavior is even more evident in Figure \ref{fig:single_view} where our method correctly predicts all 3D poses from a monocular view. In particular, notice that even the legs of the blue skeleton are predicted even if they are not visible from that particular view. (View and scene from the validation set). We suppose that the model hallucinates straight up legs since most of the people in Panoptic D2D training set are standing. 

\section{Conclusion}
We present a method for multi-person human pose estimation from calibrated views. Our neural architecture is able to predict 3D pose representations \emph{directly} from raw camera views. To the best of our knowledge, this is the first attempt to tackle such a task in a completely bottom-up fashion. The proposed method exhibits good computational scalability properties: in particular, it is essentially independent of the number of people in the scene. Moreover, it scales linearly with the number of input views.

Conducted experiments show state-of-the-art performance on the Panoptic D2D dataset as well as a good generalization on the unseen Shelf dataset.
We hope that our work can open new research lines and new scenarios. The method visibly benefits from a wide variety of configurations of people, cameras, and environments during training. Simple 3D data augmentation techniques have been explored and proven effective in enhancing the performance; however, larger datasets, both real and synthetic, could significantly increase the model capabilities.

\section*{Acknowledgment}
The authors would like to thank Igor Moiseev and all the Checkout Technologies team for the fruitful discussions and their friendly support.



\bibliographystyle{IEEEtran}
\bibliography{IEEEabrv,biblio}

\begin{thebibliography}{10}
\providecommand{\url}[1]{#1}
\csname url@samestyle\endcsname
\providecommand{\newblock}{\relax}
\providecommand{\bibinfo}[2]{#2}
\providecommand{\BIBentrySTDinterwordspacing}{\spaceskip=0pt\relax}
\providecommand{\BIBentryALTinterwordstretchfactor}{4}
\providecommand{\BIBentryALTinterwordspacing}{\spaceskip=\fontdimen2\font plus
\BIBentryALTinterwordstretchfactor\fontdimen3\font minus
  \fontdimen4\font\relax}
\providecommand{\BIBforeignlanguage}[2]{{%
\expandafter\ifx\csname l@#1\endcsname\relax
\typeout{** WARNING: IEEEtran.bst: No hyphenation pattern has been}%
\typeout{** loaded for the language `#1'. Using the pattern for}%
\typeout{** the default language instead.}%
\else
\language=\csname l@#1\endcsname
\fi
#2}}
\providecommand{\BIBdecl}{\relax}
\BIBdecl

\bibitem{dang2019deep}
Q.~Dang, J.~Yin, B.~Wang, and W.~Zheng, ``Deep learning based 2d human pose
  estimation: A survey,'' \emph{Tsinghua Science and Technology}, vol.~24,
  no.~6, pp. 663--676, 2019.

\bibitem{iskakov2019learnable}
K.~Iskakov, E.~Burkov, V.~Lempitsky, and Y.~Malkov, ``Learnable triangulation
  of human pose,'' in \emph{Proceedings of the IEEE International Conference on
  Computer Vision}, 2019, pp. 7718--7727.

\bibitem{joo2015panoptic}
H.~Joo, H.~Liu, L.~Tan, L.~Gui, B.~Nabbe, I.~Matthews, T.~Kanade, S.~Nobuhara,
  and Y.~Sheikh, ``Panoptic studio: A massively multiview system for social
  motion capture,'' in \emph{Proceedings of the IEEE International Conference
  on Computer Vision}, 2015, pp. 3334--3342.

\bibitem{cao2017realtime}
Z.~Cao, T.~Simon, S.-E. Wei, and Y.~Sheikh, ``Realtime multi-person 2d pose
  estimation using part affinity fields,'' in \emph{Proceedings of the IEEE
  Conference on Computer Vision and Pattern Recognition}, 2017, pp. 7291--7299.

\bibitem{wei2016convolutional}
S.-E. Wei, V.~Ramakrishna, T.~Kanade, and Y.~Sheikh, ``Convolutional pose
  machines,'' in \emph{Proceedings of the IEEE conference on Computer Vision
  and Pattern Recognition}, 2016, pp. 4724--4732.

\bibitem{newell2017associative}
A.~Newell, Z.~Huang, and J.~Deng, ``Associative embedding: End-to-end learning
  for joint detection and grouping,'' in \emph{Advances in neural information
  processing systems}, 2017, pp. 2277--2287.

\bibitem{papandreou2018personlab}
G.~Papandreou, T.~Zhu, L.-C. Chen, S.~Gidaris, J.~Tompson, and K.~Murphy,
  ``Personlab: Person pose estimation and instance segmentation with a
  bottom-up, part-based, geometric embedding model,'' in \emph{Proceedings of
  the European Conference on Computer Vision (ECCV)}, 2018, pp. 269--286.

\bibitem{kreiss2019pifpaf}
S.~Kreiss, L.~Bertoni, and A.~Alahi, ``Pifpaf: Composite fields for human pose
  estimation,'' in \emph{Proceedings of the IEEE Conference on Computer Vision
  and Pattern Recognition}, 2019, pp. 11\,977--11\,986.

\bibitem{feng2019}
F.~Zhang, X.~Zhu, H.~Dai, M.~Ye, and C.~Zhu, ``Distribution-aware coordinate
  representation for human pose estimation,'' 2019.

\bibitem{newell2016stacked}
A.~Newell, K.~Yang, and J.~Deng, ``Stacked hourglass networks for human pose
  estimation,'' in \emph{European conference on computer vision}.\hskip 1em
  plus 0.5em minus 0.4em\relax Springer, 2016, pp. 483--499.

\bibitem{chen2018cascaded}
Y.~Chen, Z.~Wang, Y.~Peng, Z.~Zhang, G.~Yu, and J.~Sun, ``Cascaded pyramid
  network for multi-person pose estimation,'' in \emph{Proceedings of the IEEE
  conference on computer vision and pattern recognition}, 2018, pp. 7103--7112.

\bibitem{xiao2018simple}
B.~Xiao, H.~Wu, and Y.~Wei, ``Simple baselines for human pose estimation and
  tracking,'' in \emph{Proceedings of the European conference on computer
  vision (ECCV)}, 2018, pp. 466--481.

\bibitem{ke2018multi}
L.~Ke, M.-C. Chang, H.~Qi, and S.~Lyu, ``Multi-scale structure-aware network
  for human pose estimation,'' in \emph{Proceedings of the European Conference
  on Computer Vision (ECCV)}, 2018, pp. 713--728.

\bibitem{sun2019deep}
K.~Sun, B.~Xiao, D.~Liu, and J.~Wang, ``Deep high-resolution representation
  learning for human pose estimation,'' in \emph{Proceedings of the IEEE
  Conference on Computer Vision and Pattern Recognition}, 2019, pp. 5693--5703.

\bibitem{kocabas2018multiposenet}
M.~Kocabas, S.~Karagoz, and E.~Akbas, ``Multiposenet: Fast multi-person pose
  estimation using pose residual network,'' in \emph{Proceedings of the
  European Conference on Computer Vision (ECCV)}, 2018, pp. 417--433.

\bibitem{osokin2018lightweight_openpose}
D.~Osokin, ``Real-time 2d multi-person pose estimation on cpu: Lightweight
  openpose,'' in \emph{arXiv preprint arXiv:1811.12004}, 2018.

\bibitem{zhang2019cvpr}
F.~Zhang, X.~Zhu, and M.~Ye, ``Fast human pose estimation,'' in \emph{The IEEE
  Conference on Computer Vision and Pattern Recognition (CVPR)}, June 2019.

\bibitem{tome2017lifting}
D.~Tome, C.~Russell, and L.~Agapito, ``Lifting from the deep: Convolutional 3d
  pose estimation from a single image,'' in \emph{Proceedings of the IEEE
  Conference on Computer Vision and Pattern Recognition}, 2017, pp. 2500--2509.

\bibitem{pavllo20193d}
D.~Pavllo, C.~Feichtenhofer, D.~Grangier, and M.~Auli, ``3d human pose
  estimation in video with temporal convolutions and semi-supervised
  training,'' in \emph{Proceedings of the IEEE Conference on Computer Vision
  and Pattern Recognition}, 2019, pp. 7753--7762.

\bibitem{yang20183d}
W.~Yang, W.~Ouyang, X.~Wang, J.~Ren, H.~Li, and X.~Wang, ``3d human pose
  estimation in the wild by adversarial learning,'' in \emph{Proceedings of the
  IEEE Conference on Computer Vision and Pattern Recognition}, 2018, pp.
  5255--5264.

\bibitem{sun2018integral}
X.~Sun, B.~Xiao, F.~Wei, S.~Liang, and Y.~Wei, ``Integral human pose
  regression,'' in \emph{Proceedings of the European Conference on Computer
  Vision (ECCV)}, 2018, pp. 529--545.

\bibitem{nibali20193d}
A.~Nibali, Z.~He, S.~Morgan, and L.~Prendergast, ``3d human pose estimation
  with 2d marginal heatmaps,'' in \emph{2019 IEEE Winter Conference on
  Applications of Computer Vision (WACV)}.\hskip 1em plus 0.5em minus
  0.4em\relax IEEE, 2019, pp. 1477--1485.

\bibitem{kanazawa2019learning}
A.~Kanazawa, J.~Y. Zhang, P.~Felsen, and J.~Malik, ``Learning 3d human dynamics
  from video,'' in \emph{Proceedings of the IEEE Conference on Computer Vision
  and Pattern Recognition}, 2019, pp. 5614--5623.

\bibitem{habibie2019wild}
I.~Habibie, W.~Xu, D.~Mehta, G.~Pons-Moll, and C.~Theobalt, ``In the wild human
  pose estimation using explicit 2d features and intermediate 3d
  representations,'' in \emph{Proceedings of the IEEE Conference on Computer
  Vision and Pattern Recognition}, 2019, pp. 10\,905--10\,914.

\bibitem{moon2019camera}
G.~Moon, J.~Y. Chang, and K.~M. Lee, ``Camera distance-aware top-down approach
  for 3d multi-person pose estimation from a single rgb image,'' in
  \emph{Proceedings of the IEEE International Conference on Computer Vision},
  2019, pp. 10\,133--10\,142.

\bibitem{rogez2019lcr}
G.~Rogez, P.~Weinzaepfel, and C.~Schmid, ``Lcr-net++: Multi-person 2d and 3d
  pose detection in natural images,'' \emph{IEEE transactions on pattern
  analysis and machine intelligence}, 2019.

\bibitem{bridgeman2019multi}
L.~Bridgeman, M.~Volino, J.-Y. Guillemaut, and A.~Hilton, ``Multi-person 3d
  pose estimation and tracking in sports,'' in \emph{Proceedings of the IEEE
  Conference on Computer Vision and Pattern Recognition Workshops}, 2019, pp.
  0--0.

\bibitem{ohashi2020synergetic}
T.~Ohashi, Y.~Ikegami, and Y.~Nakamura, ``Synergetic reconstruction from 2d
  pose and 3d motion for wide-space multi-person video motion capture in the
  wild,'' \emph{arXiv preprint arXiv:2001.05613}, 2020.

\bibitem{martinez2017simple}
J.~Martinez, R.~Hossain, J.~Romero, and J.~J. Little, ``A simple yet effective
  baseline for 3d human pose estimation,'' in \emph{Proceedings of the IEEE
  International Conference on Computer Vision}, 2017, pp. 2640--2649.

\bibitem{pavlakos2017harvesting}
G.~Pavlakos, X.~Zhou, K.~G. Derpanis, and K.~Daniilidis, ``Harvesting multiple
  views for marker-less 3d human pose annotations,'' in \emph{Proceedings of
  the IEEE conference on computer vision and pattern recognition}, 2017, pp.
  6988--6997.

\bibitem{tome2018rethinking}
D.~Tome, M.~Toso, L.~Agapito, and C.~Russell, ``Rethinking pose in 3d:
  Multi-stage refinement and recovery for markerless motion capture,'' in
  \emph{2018 international conference on 3D vision (3DV)}.\hskip 1em plus 0.5em
  minus 0.4em\relax IEEE, 2018, pp. 474--483.

\bibitem{qiu2019cross}
H.~Qiu, C.~Wang, J.~Wang, N.~Wang, and W.~Zeng, ``Cross view fusion for 3d
  human pose estimation,'' in \emph{Proceedings of the IEEE International
  Conference on Computer Vision}, 2019, pp. 4342--4351.

\bibitem{pirinen2019domes}
A.~Pirinen, E.~G{\"a}rtner, and C.~Sminchisescu, ``Domes to drones:
  Self-supervised active triangulation for 3d human pose reconstruction,'' in
  \emph{Advances in Neural Information Processing Systems}, 2019, pp.
  3907--3917.

\bibitem{belagiannis20143d}
V.~Belagiannis, S.~Amin, M.~Andriluka, B.~Schiele, N.~Navab, and S.~Ilic, ``3d
  pictorial structures for multiple human pose estimation,'' in
  \emph{Proceedings of the IEEE Conference on Computer Vision and Pattern
  Recognition}, 2014, pp. 1669--1676.

\bibitem{tanke2019iterative}
J.~Tanke and J.~Gall, ``Iterative greedy matching for 3d human pose tracking
  from multiple views,'' in \emph{German Conference on Pattern
  Recognition}.\hskip 1em plus 0.5em minus 0.4em\relax Springer, 2019, pp.
  537--550.

\bibitem{dong2019fast}
J.~Dong, W.~Jiang, Q.~Huang, H.~Bao, and X.~Zhou, ``Fast and robust
  multi-person 3d pose estimation from multiple views,'' in \emph{Proceedings
  of the IEEE Conference on Computer Vision and Pattern Recognition}, 2019, pp.
  7792--7801.

\bibitem{howard2017mobilenets}
A.~G. Howard, M.~Zhu, B.~Chen, D.~Kalenichenko, W.~Wang, T.~Weyand,
  M.~Andreetto, and H.~Adam, ``Mobilenets: Efficient convolutional neural
  networks for mobile vision applications,'' \emph{arXiv preprint
  arXiv:1704.04861}, 2017.

\bibitem{jaderberg2015spatial}
M.~Jaderberg, K.~Simonyan, A.~Zisserman \emph{et~al.}, ``Spatial transformer
  networks,'' in \emph{Advances in neural information processing systems},
  2015, pp. 2017--2025.

\bibitem{moon2018v2v}
G.~Moon, J.~Yong~Chang, and K.~Mu~Lee, ``V2v-posenet: Voxel-to-voxel prediction
  network for accurate 3d hand and human pose estimation from a single depth
  map,'' in \emph{Proceedings of the IEEE conference on computer vision and
  pattern Recognition}, 2018, pp. 5079--5088.

\bibitem{belagiannis2014multiple}
V.~Belagiannis, X.~Wang, B.~Schiele, P.~Fua, S.~Ilic, and N.~Navab, ``Multiple
  human pose estimation with temporally consistent 3d pictorial structures,''
  in \emph{European Conference on Computer Vision}.\hskip 1em plus 0.5em minus
  0.4em\relax Springer, 2014, pp. 742--754.

\bibitem{belagiannis20153d}
V.~Belagiannis, S.~Amin, M.~Andriluka, B.~Schiele, N.~Navab, and S.~Ilic, ``3d
  pictorial structures revisited: Multiple human pose estimation,'' \emph{IEEE
  transactions on pattern analysis and machine intelligence}, vol.~38, no.~10,
  pp. 1929--1942, 2015.

\bibitem{ershadi2018multiple}
S.~Ershadi-Nasab, E.~Noury, S.~Kasaei, and E.~Sanaei, ``Multiple human 3d pose
  estimation from multiview images,'' \emph{Multimedia Tools and Applications},
  vol.~77, no.~12, pp. 15\,573--15\,601, 2018.

\bibitem{li2017light}
Z.~Li, C.~Peng, G.~Yu, X.~Zhang, Y.~Deng, and J.~Sun, ``Light-head r-cnn: In
  defense of two-stage object detector,'' \emph{arXiv preprint
  arXiv:1711.07264}, 2017.

\end{thebibliography}
%


\end{document}